\pdfoutput=1
\documentclass[letterpaper, 10 pt, conference]{ieeeconf}  % Comment this line out if you need a4paper

\IEEEoverridecommandlockouts                              % This command is only needed if
\usepackage{color}
\definecolor{green}{rgb}{0, 0.5, 0}
\definecolor{orange}{rgb}{0.8, 0.6, 0.2}
\definecolor{red}{rgb}{1.0, 0.0, 0.0}
\definecolor{teal}{rgb}{0.0, 0.4, 0.4}
\definecolor{purple}{rgb}{0.65,0,0.65}
\definecolor{saffron}{rgb}{0.95,0.75,0.2}
\definecolor{turquoise}{rgb}{0.0,0.5,0.5}
\definecolor {mygray}{gray}{.9}

\usepackage[normalem]{ulem}

\usepackage{amsmath}

\def\*#1{\mathbf{#1}}

\usepackage{hyperref}

\usepackage{overpic}
\usepackage{subfig}

\usepackage{algorithm}
\usepackage{algpseudocode}
\usepackage{pifont}
\usepackage{algorithmicx}

\usepackage{booktabs}
\usepackage{colortbl}
\usepackage{makecell}
\usepackage{multirow}

\let\labelindent\relax
\usepackage[inline]{enumitem}

\usepackage{graphicx}

\title{\LARGE \bf
Robust and Precise Vehicle Localization based on Multi-sensor Fusion in Diverse City Scenes
}

\author{Guowei Wan, Xiaolong Yang, Renlan Cai, Hao Li, Hao Wang, Shiyu Song$^{1}$% <-this % stops a space
\thanks{*This work is supported by Baidu Autonomous Driving Business Unit in conjunction with the \href{http://apollo.auto/}{Apollo Project}.}% <-this % stops a space
\thanks{The authors are with Baidu Autonomous Driving Business Unit, \texttt{\{wanguowei, yangxiaolong02, cairenlan, lihao30, wanghao29, songshiyu\}@baidu.com.}}%
\thanks{$^{1}$Author to whom correspondence should be addressed, E-mail: {\tt\small \href{mailto:songshiyu@baidu.com}{songshiyu@baidu.com}}}%
}

\begin{document}

\maketitle
\thispagestyle{empty}
\pagestyle{empty}

\begin{abstract}

We present a robust and precise localization system that achieves centimeter-level localization accuracy in disparate city scenes. Our system adaptively uses information from complementary sensors such as GNSS, LiDAR, and IMU to achieve high localization accuracy and resilience in challenging scenes, such as urban downtown, highways, and tunnels. Rather than relying only on LiDAR intensity or 3D geometry, we make innovative use of LiDAR intensity and altitude cues to significantly improve localization system accuracy and robustness. Our GNSS RTK module utilizes the help of the multi-sensor fusion framework and achieves a better ambiguity resolution success rate. An error-state Kalman filter is applied to fuse the localization measurements from different sources with novel uncertainty estimation. We validate, in detail, the effectiveness of our approaches, achieving 5-10cm RMS accuracy and outperforming previous state-of-the-art systems. Importantly, our system, while deployed in a large autonomous driving fleet, made our vehicles fully autonomous in crowded city streets despite road construction that occurred from time to time. A dataset including more than 60 km real traffic driving in various urban roads is used to comprehensively test our system.
%Both quantitative and qualitative analysis has been carried out.

\end{abstract}

\section{Introduction}
Vehicle localization is one of the fundamental tasks in autonomous driving. 
The single-point positioning accuracy of the global navigation satellite system (GNSS) is about 10m due to satellite orbit and clock errors, together with tropospheric and ionospheric delays. These errors can be calibrated out with observations from a surveyed reference station. The carrier-phase based differential GNSS technique, known as Real Time Kinematic (RTK), can provide centimeter positioning accuracy \cite{groves2013principles}.
The most significant advantage of RTK is that it provides almost all-weather availability. However, its disadvantage is equally obvious that it's highly vulnerable to signal blockage, multi-path because it relies on the precision carrier-phase positioning techniques.
%Besides, it is a differential global navigation satellite system (DGNSS). An RTK reference station that broadcasts differential signals to running vehicles, sometimes known as rovers, is required for the DGNSS. Maintaining a reliable signal transmission connection for a running vehicle is difficult in practice even via the modern 3G/4G wireless communication network.
Intuitively, LiDAR is a promising sensor for precise localization. 
%In recent years, there has been enormous interest in LiDAR-based localization \cite{Levinson_etal2007, kummerle2008monte, kummerle2009autonomous, Levinson_etal2010, baldwin2012road, baldwin2012laser, sheehan2013continuous, yoneda2014lidar, maddern2015leveraging, wolcott2015fast, Wolcott_etal2017}. We fully acknowledge the advantages of the LiDAR sensor, and propose an improved LiDAR based localization method in this work.
Failure during harsh weather conditions and road construction still is an important issue of LiDAR-based methods, although related works have shown good progress in solving these problems, for example, light rain \cite{Levinson_etal2007} and snow \cite{wolcott2015fast}. Furthermore, LiDAR and RTK are two sensors those are complementary in terms of applicable scenes. LiDAR works well when the environment is full of 3D or texture features,  while RTK performs excellently in open space.
An inertial measurement unit (IMU), including the gyroscopes and the accelerometers, continuously calculate the position, orientation, and velocity via the technology that is commonly referred to the dead reckoning. It's self-contained navigation method, that is immune to jamming and deception. But it suffers badly from integration drift.
%For a tactical grade MEMS IMU, it typically drifts more than 1m within 10 seconds without the aid of other positioning methods.

\begin{figure}[!!t]
	\centering
	\includegraphics[width=0.99\columnwidth]{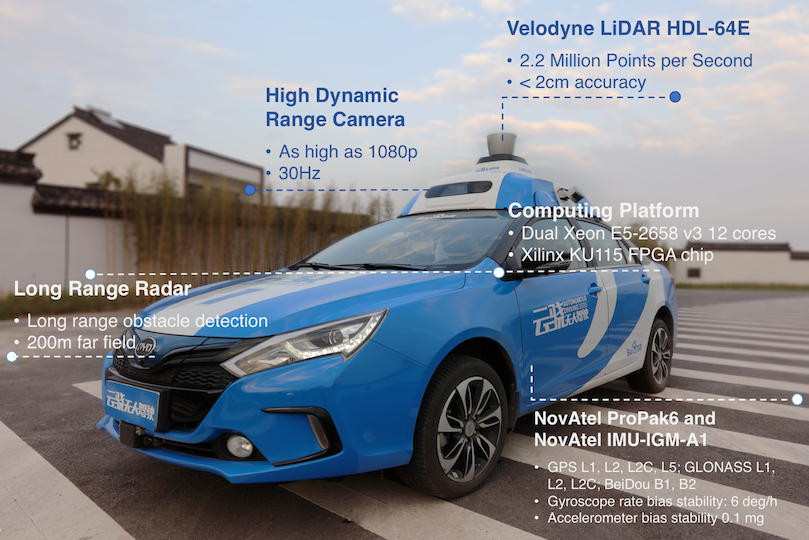}
	\caption{
		\footnotesize Our autonomous vehicle is equipped with a Velodyne LiDAR HDL-64E. An integrated navigation system, NovAtel ProPak6 plus NovAtel IMU-IGM-A1, is installed for raw sensor data collection, such as GNSS pseudo range and carrier wave, IMU specific force and rotation rate. A computing platform equipped with Dual Xeon E5-2658 v3 12 cores, and a Xilinx KU115 FPGA chip with 55\% utilization for LiDAR localization.
		%The built-in tightly integrated inertial and satellite navigation output was not used. This is different from previous works \cite{Levinson_etal2007,Levinson_etal2010} that rely on the inputs from the GNSS-IMU integrated navigation equipment.
		%Long range radars and cameras with high dynamic range are also installed, but they were not used for localization.
	}
	\label{fig:intro}
	\vspace{-0.2cm}
\end{figure}

\begin{figure*}[!!t]
	\centering
	\includegraphics[width=1.75\columnwidth]{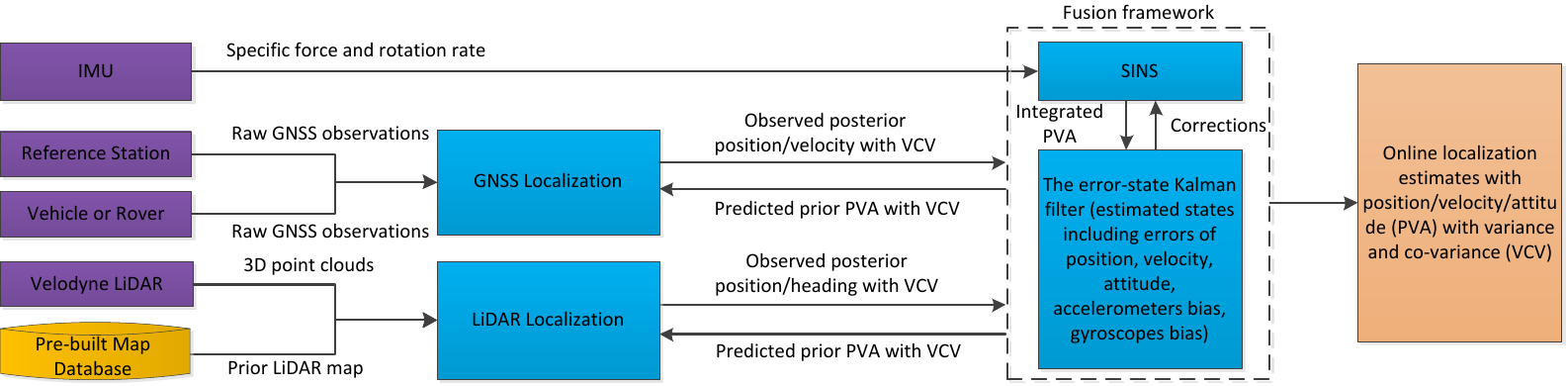}
	\caption{
		\footnotesize Overview of the architecture of our system that estimates the optimal position, velocity, attitude (PVA) of the autonomous vehicle by combining sensor input (purple) with pre-built LiDAR map (yellow). GNSS and LiDAR estimate the PVA used by an error-state Kalman filter as the measurements, while the Kalman filter provides the predicted prior PVA. The strap-down inertial navigation system (SINS) is used as a prediction model in the Kalman filter propagation phase by integrating the specific force  $\boldsymbol{\mathit{f}}^b$ measured by the accelerometer and the rotation rate $\boldsymbol{\mathit{\omega}}_{ib}^b$ measured by the gyroscope. The corrections including the bias of accelerometer and gyroscope, the errors of PVA, etc estimated by the Kalman filter are fed to the SINS.
	}
	\label{fig:overview}
	\vspace{-0.2cm}
\end{figure*}

Thus, each sensor has its own unique characteristics and its working conditions. Here, we propose a robust and precise localization system using multi-sensor fusion designed for autonomous vehicles driving in complex urban and highway scenes. More precisely, we adaptively fuse different localization methods based on sensors such as LiDAR, RTK, and IMU. The sensor configuration of our system is shown in Figure \ref{fig:intro}. Our system provides stable, resilient and precise localization service to other modules in an autonomous vehicle, which has the capability of driving in several complex scenes, such as downtown, tunnels, tree-lined roads, parking garages, and highways. We demonstrate large-scale localization using over 60 km of data in dynamic urban and highway scenes.
%We discuss LiDAR based localization in detail in Section \ref{section:lidar}. RTK based localization is discussed in Section \ref{section:gnss}. How the inertial sensor is introduced into the fusion framework is shown in Section \ref{section:ins}. The overall sensor fusion framework based on EKF is introduced in Section \ref{section:fusion}.
In Figure \ref{fig:overview}, we show the architecture of our multi-sensor fusion framework.

To summarize, our main contributes are:
\begin{itemize}
	\item A joint framework for vehicle localization that adaptively fuses different sensors including LiDAR, RTK, and IMU. It effectively leverages their advantages and shields our system from their failure in various scenes by having effective uncertainty estimation.
	\item A LiDAR localization method that adaptively combines intensity and altitude cues to achieve robust and accurate results especially during challenging situations like road construction, outperforming previous works.
	\item A vehicle localization system that has been rigorously tested daily in crowded urban streets, making our vehicles fully autonomous in various challenging scenes including urban downtown, highways, and tunnels.
\end{itemize}

\section{Related Work}
\label{section:related}
Multi-sensor fusion is not a brand new idea. However, fusing multiple sensors and making the whole system accurate, robust and applicable for various scenes is a very challenging task. A. Soloviev. \cite{soloviev2008tight} and Y. Gao et al. \cite{gao2015ins} implemented integrated GNSS/LiDAR/IMU navigation systems using a 2D laser scanner plus GNSS and IMU. The applicable scenes are limited due to the LiDAR localization module that relies on particular features such as building walls. LiDAR aided by inertial sensors can also localize the autonomous vehicle. It provides the localization measurements, while inertial sensors typically are used to predict the incremental movement between scans to improve the point cloud matching. These works \cite{soloviev2007tight, tang2015lidar, hemann2016long} are among them. Their methods rely on LiDAR solely, but RTK is a perfect complementary localization method to LiDAR. RTK plays an important role, especially in open spaces or streets with road construction.

LiDAR-based vehicle localization has been very popular in recent years. R. K\"{u}mmerle and W. Burgard \cite{kummerle2009autonomous} developed an autonomous system that utilizes multi-level surface maps of corresponding environments to localize itself based on the particle filter. K. Yoneda and S. Mita \cite{yoneda2014lidar} localized their autonomous vehicle by using Iterative Closest Point (ICP) method \cite{besl1992method} to align the real-time scanning point cloud to the geo-referenced map. However, it is known that methods like ICP are very sensitive to the initial guess. They can fail in scenes without abundant 3D features, such as highways or other open spaces. The works that are closest to ours are from J. Levinson and S. Thrun \cite{Levinson_etal2007, Levinson_etal2010}. They propose a LiDAR intensity-based localization method. LiDAR intensity provides more texture information of the environment as valuable additional cues compared to the localization system that is based solely on the 3D geometry of the point cloud. We improved the methodology for various aspects, including a new image alignment step to refine the heading angle estimation, a joint cost function involving both the intensity and the altitude to achieve robust results, and a new step to estimate the covariance matrices of the result. Methods mentioned thus far are designed for multiple layers LiDAR, such as Velodyne HDL-64E, Velodyne HDL-32E or even Velodyne VLP-16. It is expected that the retail price of these LiDAR scanners will fall quickly because there are more than 40 manufacturers competing in this rising field as we are writing this article. Rather than using multi-layer LiDARs, works \cite{adams2004particle, kummerle2008monte, baldwin2012road, baldwin2012laser, sheehan2013continuous, chong2013synthetic, maddern2015leveraging} attempt to accomplish the similar task with 2D or low-end LiDARs.
\section{LiDAR Map Generation}
\label{section:map}
\begin{figure}[htbp]
	\centering
	\begin{minipage}[b]{0.45\linewidth}	
		\subfloat[]{
			\centering
			\includegraphics[width=2.35in]{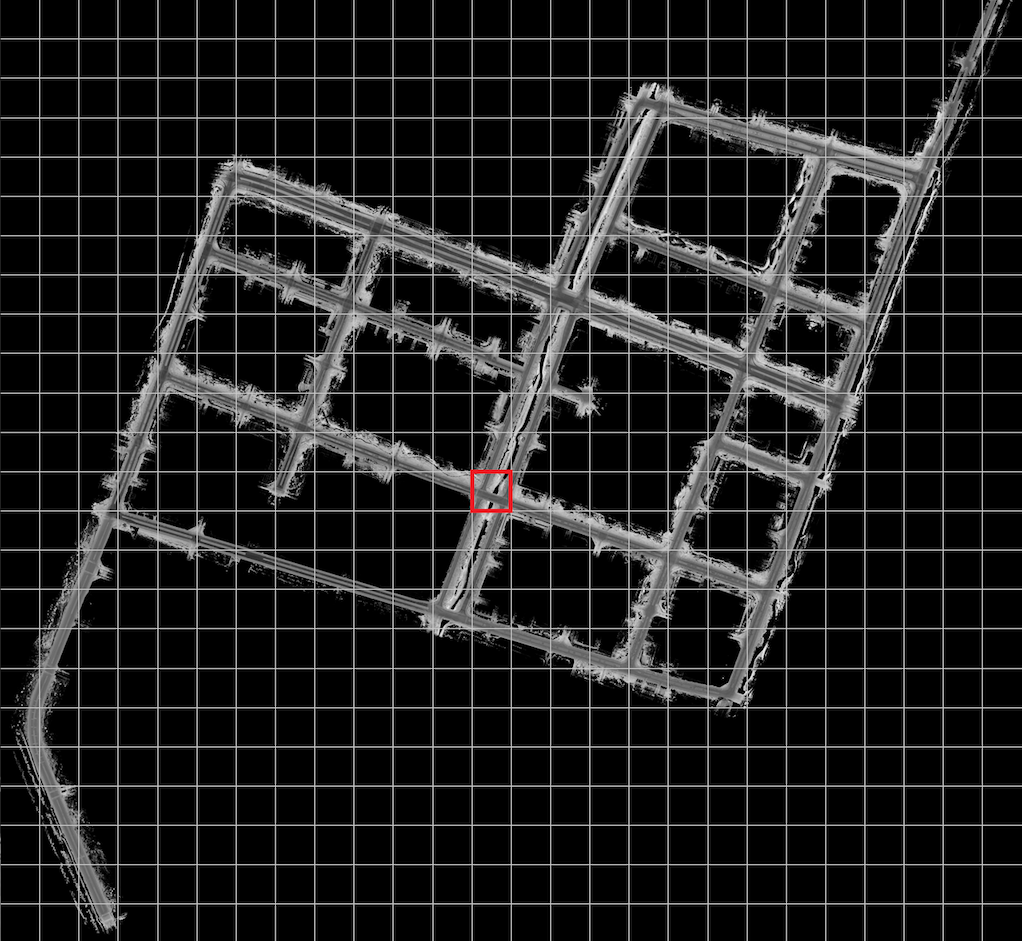}
		}
	\end{minipage}	
	\hfill
	\begin{minipage}[b]{0.28\linewidth}		
		\subfloat[]{
			\centering
			\includegraphics[width=0.92in]{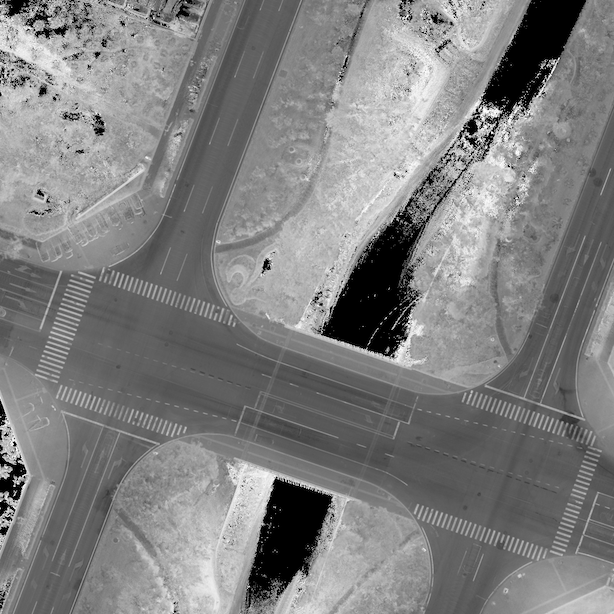} 
		} \\
		%\vfill
		\subfloat[]{
			\centering
			\includegraphics[width=0.92in]{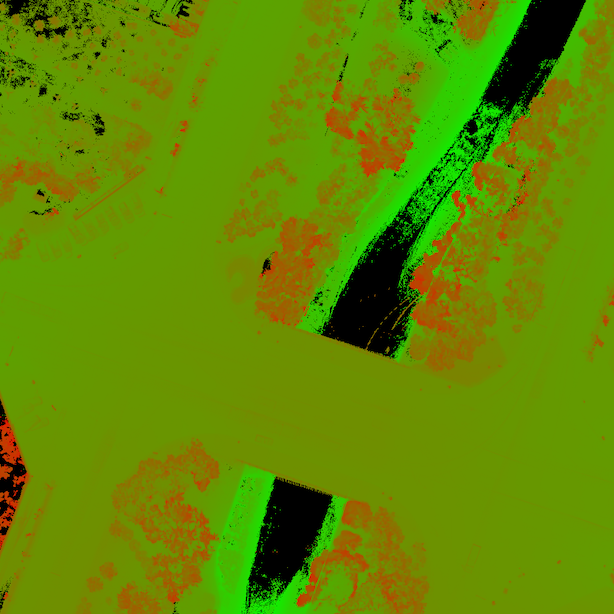} 
		}
	\end{minipage}	
	\caption{\footnotesize An example of a pre-built LiDAR map including the statistics of laser intensity and altitude. (a) A pre-built map composed covering 3.3km $\times$ 3.1km area rendered in laser intensity values. (b) A zoomed-in view of (a). (c) The same zoomed-in view of (a), but rendered in altitude $a$.}
	\label{fig:map}
	\vspace{-0.2cm}
\end{figure}

Our LiDAR based localization module relies on a pre-generated map. Our goal here is to obtain a grid-cell representation of the environment. Each cell stores the statistics of laser reflection intensity and altitude. In \cite{Levinson_etal2007} and \cite{Levinson_etal2010}, each cell is represented by a single Gaussian distribution model which maintains the average and the variance of the intensity. In \cite{Wolcott_etal2017}, this is extended to a Gaussian mixture model (GMM) maintaining both the intensity and the altitude. In our work, we still use single Gaussian distribution to model the environment but involve both the intensity and the altitude as shown in Figure \ref{fig:map}. We found that combining both the intensity and the altitude measurements in the cost function through an adaptive weighting method can significantly improve the localization accuracy and robustness. This will be discussed in details in Section \ref{lidar:horizontal} and \ref{section:lidarexperiments}.

%The map is built by collecting LiDAR scans, GNSS and IMU logs on a vehicle equipped with similar sensors to our autonomous driving vehicles as shown in Figure \ref{fig:intro}. In open spaces with good GNSS signal reception, the GNSS/INS solution based on post-processing algorithms, such as NovAtel Inertial Explorer \cite{NovAtelIE}, is able to produce enough accurate vehicle motion trajectories. In complex urban environments, we treat it as a classic SLAM problem with loop closure and global pose-graph optimization. Finally, We project all of our 3D LiDAR scan points onto the ground plane and generate the statistics of the intensity and the altitude. The resolution of a cell in our map is 12.5cm. 

\section{LiDAR based Localization}
\label{section:lidar}
Our localization system estimates the position, velocity and attitude (PVA) jointly. However, we seek an optimal solution that includes only the 3-dimensional position and the heading, $(x, y, a, h)$ in the LiDAR localization module. Here $(x, y)$ is a 2D Cartesian coordinate from a conformal projection, such as the Universal Transverse Mercator (UTM). Our pre-built map includes the statistics of the altitude of the road surface. We could obtain the altitude estimation $a$ by assuming the vehicle runs on the surface of the road. \cite{Wolcott_etal2017} and \cite{Olson2009} make exhaustive searching over $(x, y, h)$, which causes the computational complexity of $O(n^3)$. For better efficiency, we have a separate $h$ estimation step based on the Lucas-Kanade algorithm \cite{LucasKanade1981} and a histogram filter for the horizontal $(x, y)$ estimation. 
Please refer to Algorithm~\ref{alg:lidar} for a complete step-by-step overview.

\begin{algorithm}
	\begin{small}
	\caption{LiDAR-based localization}
	\label{alg:lidar}
	\begin{algorithmic}[1] % The number tells where the line numbering should start
		\Require{Prior map $m$, online point cloud $z$, rough transformation $T_0 = (x_0, y_0, a_0, \phi_0, \theta_0, h_0)$ and search space $X$, $Y$.}
		\Ensure{Best registration ($\hat{x}, \hat{y}, \hat{a}, \hat{h}$), and covariance matrix $\mathcal{C}_{xy}$.}
		% heading angle estimation
		\State $\hat{h}\gets$ heading angle estimation \Comment{~\ref{lidar:heading}}
		\State $\hat{a_0}\gets m(x_0, y_0)$ \Comment{Get altitude from map}
		% generate map node for online point cloud
		\State Transform $z$ with the transformation $(x_0, y_0, \hat{a_0}, \phi_0, \theta_0, \hat{h})$
		%\State $project \ z \ to \ a \ map \ node \ z' \ by \ (x_0, y_0, \hat{a}, \phi_0, \theta_0, \hat{h})$
		\For{$x_i, y_i \in \{x_0 + X, y_0 + Y\} $}
		%\State $P_r\gets P(z_r|x_i, y_i, m)$  \Comment{~\ref{lidar:measure}}
		%\State $P_a\gets P(z_a|x_i, y_i, m)$  \Comment{~\ref{lidar:measure}}		
		\State $P_r \gets SSD_r(x_i,y_i,z,m)$ \Comment{Equ.~\ref{equ:ssdr}~\ref{equ:measure3}}
		\State $P_a \gets SSD_a(x_i,y_i,z,m)$ \Comment{Equ.~\ref{equ:ssda}~\ref{equ:measure3}}
		\State $P(z|x_i,y_i,m)\gets (P_r)^\gamma \cdot (P_a)^{1-\gamma}$  \Comment{Equ.~\ref{equ:measure1}~\ref{equ:measure4}~\ref{equ:measure5}}
		\State $P(x_i, y_i) \gets P(z|x_i,y_i,m) \cdot (\bar{P}(x_i,y_i))^{1 / \kappa}$ \Comment{Equ.~\ref{equ:poster}}
		\EndFor
		\State $(\hat{x}, \hat{y}) \gets \{P(x_i, y_i)\}$ \Comment{Equ.~\ref{equ:histogram1}}
		\State $\mathcal{C}_{xy} \gets \{P(x_i, y_i)\}$ \Comment{Equ.~\ref{equ:histogram2}}
		%\State $\bar{P}(x,y) \gets \{P(x_i, y_i)\}$ \Comment{Equ.~\ref{equ:motion}}
		\State $\hat{a}\gets m(\hat{x}, \hat{y})$ \Comment{Get altitude from map}
		\State \textbf{return} $(\hat{x}, \hat{y}, \hat{a}, \hat{h}, \mathcal{C}_{xy})$ 
	\end{algorithmic}
	\end{small}
\end{algorithm}

\vspace{-0.2cm}
\subsection{Heading Angle Estimation}
\label{lidar:heading}
Similar to the procedures in the map generation section, we again project the online point cloud onto the ground plane, and an intensity image is generated similar to Figure \ref{fig:map}, but the pixels filled with the intensity values in the image are thinly dispersed due to the sparsity of the online point cloud. We can obtain the localization result $(x, y, h)$ by matching this sparse image with the pre-built LiDAR map. During the heading angle $h$ estimation step, Lucas-Kanade algorithm \cite{LucasKanade1981} is applied, which is a technique that uses the image gradient to search for the best match between two images. More precisely, a type of Lucas-Kanade algorithm, the forwards additive algorithm \cite{baker2004lucas}, is applied. It is essentially a Gauss-Newton gradient descent non-linear optimization algorithm. We verified that the Lucas-Kanade algorithm can converge to solutions with sub-degree accuracy using the online intensity image as the template image. A tactical grade MEMS IMU cannot provide sufficiently accurate heading angle estimation by itself. The effectiveness of this step is shown in Table \ref{table:tab3} in Section \ref{section:lidarexperiments}. Although the Lucas-Kanade algorithm can produce the horizontal translation estimation, we found that it is not accurate and robust enough in practice. Therefore, we use only the rotation component as the heading angle estimation, which is used in the next step, the horizontal localization.

\vspace{-0.2cm}
\subsection{Horizontal Localization}
\label{lidar:horizontal}
Histogram filter is one of nonparametric approaches. It approximates the posteriors by decomposing the state space into finitely many regions, and representing the cumulative posterior for each region by a single probability value. For completeness, we quote the equations of the prediction and update phase of the histogram filter from \cite{thrun2005probabilistic}:
\begin{small}
	\begin{align} 
		& \bar{P}_{k,t}=\sum_{i}P(X_t=x_k | u_t, X_{t-1}=x_i) \cdot P_{i,t-1} \label{equ:hf:pre}	\\
		& P_{k,t} = \eta \cdot P(z_t | X_t = x_k) \cdot \bar{P}_{k,t} \label{equ:hf:poster},
	\end{align}
\end{small}
where $P_{k,t}$ represents the belief of each state $x_k$ at time $t$, $u_t$ is the control input, and $z_t$ is the measurement vector.

We apply the histogram filter to the horizontal localization. The state contains $(x, y)$. The reason we use the histogram filter is that the cumulative probabilities of the posterior density are calculated in a discrete way. This exhaustive search ensures the optimal solution. Equations~\ref{equ:hf:pre} and ~\ref{equ:hf:poster} correspond to the prediction and update step, and the corresponding equations of our method are Equations~\ref{equ:motion} and ~\ref{equ:poster}, respectively.

\vspace{0.2em}
\subsubsection{The Prediction Step}
\label{lidar:motion}
The prediction step is to predict on the new belief of the state with the historical distribution of the filter and the control variables or the motion model. Here we directly update the histogram filter center using the motion prediction from the SINS, and update the belief distribution with a random walk with Gaussian noise. Previous works \cite{Levinson_etal2007, Levinson_etal2010} rely on a pre-fused GNSS/IMU solution. A complete multi-sensor fusion framework allows our system to work under varied challenging circumstances. Thus, the prediction step updates the probability of each cell as follows:
\begin{small}
	\begin{equation} \label{equ:motion}
	\bar{P}(x,y) = \eta \cdot \sum_{i,j}{P(i,j) \cdot \exp(-\frac{(i-x)^{2}+(j-y)^{2}}{2\sigma^2})},
	\end{equation}
\end{small}
where $\bar{P}(x,y)$ is the predicted probability, $P(i,j)$ is the probability after the motion update from the SINS. Here $\sigma$ is the parameter describing the rate of drift between two frames.

\vspace{0.2em}
\subsubsection{The Update Step}
\label{lidar:measure}
The second step of the histogram filter is the measurement update step, in which the posterior belief of each state is estimated by the following equation:
\begin{small}
	\begin{equation} \label{equ:poster}
	P(x,y|z,m) = \eta \cdot P(z|x,y,m) \cdot (\bar{P}(x,y))^{1 / \kappa},
	\end{equation}
\end{small}
where $z$ is the online map built with the online laser scans in the way identical  to the mapping procedure in Section \ref{section:map}, $N_z$ is the number of the cells of the online map with valid data. $m$ is the pre-built LiDAR map, $(x, y)$ is the motion pose, $\eta$ is the normalizing constant, and $\kappa$ is the Kullback-Leibler divergence \cite{kullback1951} between the distributions of $\sum_{xy}P(z|x,y,m)$ and $\sum_{xy}\bar{P}(x,y)$. The KL divergence depicts the discrimination information and is used to balance the influence of the prediction. Unlike \cite{Levinson_etal2010}, we discard the uncertainty of the GNSS/IMU pose. We have a more comprehensive fusion framework fusing the input from various sources including GNSS.

The likelihood $P(z|x,y,m)$ is calculated by matching the online point cloud with the pre-built map. We adaptively fuse the intensity and the altitude measurement by building a cost function with a dynamic weighting parameter $\gamma$:
\begin{equation} \label{equ:measure1}
P(z|x,y,m) = \eta \cdot P(z_r|x,y,m)^\gamma \cdot P(z_a|x,y,m)^{1-\gamma},
\end{equation}
where $z_r$ and $z_a$ represent the intensity and altitude measurement of the online sensor input.

The intensity part is formalized as below:
\begin{footnotesize}
	\begin{equation} \label{equ:ssdr}
	SSD_r = \sum_{i,j} \frac{(r_{m_{(i-x,j-y)}}-r_{z_{(i,j)}})^{2} (\sigma_{m_{(i-x,j-y)}}^{2} + \sigma_{z_{(i,j)}}^{2})}{\sigma_{m_{(i-x,j-y)}}^{2} \sigma_{z_{(i,j)}}^{2}},
	\end{equation}
\end{footnotesize}
where $r_m$, and $r_z$ denote the average intensity value in the pre-built map and the online map, respectively. $\sigma_{m}$ and $\sigma_{z}$ represent the standard deviation of the intensity value. SSD (Sum of Squared Differences) is used to evaluate the similarity between the sensor input and the map. The impact of the environment change is implicitly diminished by the variance term. Two significantly different intensity values with both low variance values imply the environment change.

Similarly, the SSD of altitude is taken into account as below:
\begin{footnotesize}
	\begin{equation} \label{equ:ssda}
	SSD_a = \sum_{i,j} (a_{m_{(i-x,j-y)}}-a_{z_{(i,j)}})^{2},
	\end{equation}
\end{footnotesize}
where $a_m$ and $a_z$ represent the average altitude of the pre-built map and the online map, respectively. We remove the variance term here in $SSD_a$ because the altitude is variable vertically by nature.

The likehood of $P(z_r|x,y,m)$ and $P(z_a|x,y,m)$ are defined as:
\begin{footnotesize}
	\begin{equation} \label{equ:measure3}
	P(z_r|x,y,m) = \eta \cdot \alpha^{-\frac{SSD_r}{2 \cdot N_z}}, \ P(z_a|x,y,m) = \eta \cdot \alpha^{-\frac{\lambda \cdot SSD_a}{2 \cdot N_z}},
	\end{equation}
\end{footnotesize}
when the parameter $\alpha = e$, they are Gaussian-like probability distribution. We change $\alpha$ to adjust its smoothness.

The adaptive weight parameter $\gamma$ plays an important role during the intensity and altitude fusion. We let it determined by the variances of $[P(z_r|x,y,m)]_{xy}$ and $[P(z_a|x,y,m)]_{xy}$. The variances are defined by the following equations:
\begin{footnotesize}
	\begin{equation} \label{equ:measure4}
	\begin {split}
	\sigma_{x}^{2} = \frac{\sum_{xy} P(x,y)^\beta (x-\bar{x})^2 }{\sum_{xy} P(x,y)^\beta}, \
	\sigma_{y}^{2} = \frac{\sum_{xy} P(x,y)^\beta (y-\bar{y})^2 }{\sum_{xy} P(x,y)^\beta},
	\end{split}
	\end{equation}
\end{footnotesize}
where $\bar{x}$ and $\bar{y}$ are the center of mass of distribution.

Therefore, the $x$ and $y$ variances of intensity and altitude are represented as $\sigma_{x}^{2}(r)$, $\sigma_{y}^{2}(r)$, $\sigma_{x}^{2}(a)$, and $\sigma_{y}^{2}(a)$. The weight $\gamma$ is computed as:
\begin{small}
	\begin{equation} \label{equ:measure5}
	\gamma = \frac{\sigma_{x}^{2}(a) \sigma_{y}^{2}(a)}{\sigma_{x}^{2}(a) \sigma_{y}^{2}(a) + \sigma_{x}^{2}(r) \sigma_{y}^{2}(r)}.
	\end{equation}
\end{small}

\vspace{0.2em}
\subsubsection{Optimal Offset}
\label{lidar:offset}
%Once we obtain the final posterior distribution, the $x$ and $y$ offsets are estimated by the center of the mass of the posterior distribution. Instead of using all the cells of the histogram filter to calculate the optimal offset, we use only a small square of cells around the chosen cell. The chosen cell is the cell that with the largest value $P(x_i,y_i)$ or the second largest value $P(x_j,y_j)$. If $P(x_i,y_i) / P(x_j,y_j)$ is less than a given threshold (we use $1.1$) and the distance of $i$-th cell and the center of the filter is larger than $j$-th cell's, $j$-th cell will be the chosen cell. The final $x$ and $y$ offsets are calculated by the following equations:
%\begin{small}
%\begin{equation} \label{equ:histogram1}
%	\hat{x} = \frac{\sum_{x,y} P(x,y)^\beta \cdot x}{\sum_{x,y}P(x,y)^\beta} \qquad \hat{y} = \frac{\sum_{x,y} P(x,y)^\beta \cdot y}{\sum_{x,y}P(x,y)^\beta}
%\end{equation}
%\end{small}
The optimal offset is estimated from the posterior distribution of the histogram filter. Instead of using all the states of the histogram filter to calculate the optimal offset, we use only a small squared area around the state of the largest or the second largest posterior belief. If the value of the second largest posterior belief achieves a given ratio of the largest one and is closer to the center of the histogram filter, we take the state of the second largest posterior belief as the center of the small squared area. Otherwise, we take the state of the largest posterior belief. Assuming the small area is $\mathcal{Z}$, the optimal offset $(\hat{x}, \hat{y})$ is calculated by:
\begin{footnotesize}
	\begin{equation} \label{equ:histogram1}
	\hat{x} = \frac{\sum_{(x,y) \in \mathcal{Z}} P(x,y)^\beta \cdot x}{\sum_{(x,y) \in \mathcal{Z}}P(x,y)^\beta}, \ \hat{y} = \frac{\sum_{(x,y) \in \mathcal{Z}} P(x,y)^\beta \cdot y}{\sum_{(x,y) \in \mathcal{Z}}P(x,y)^\beta}.
	\end{equation}
\end{footnotesize}

\vspace{0.2em}
\subsubsection{Uncertainty Estimation}
\label{lidar:uncertainty}
The localization result is used to update the Kalman filter in section~\ref{section:fusion}. The key issue is the evaluation of the uncertainty associated with the state estimates. We let the vector $\vec{t}_d=(\hat{x},\hat{y})^T$ and $\vec{t}_{x,y} = (x,y)^T$ be the optimized offset and the offset of the $(x,y)$ cell, respectively. The resulting covariance matrix $\mathcal{C}_{xy}$ can be computed as:
\begin{small}
	\begin{equation} \label{equ:histogram2}
	\mathcal{C}_{xy} = \frac{1}{\sum_{x,y}P(x,y)^\beta} \cdot \sum_{xy} P(x,y)^\beta \cdot (\vec{t}_{x,y} - \vec{t}_d)(\vec{t}_{x,y} - \vec{t}_d)^T.
	\end{equation}
\end{small}
In Figure~\ref{fig:exp1b} from the experimental section, we show samples of the filter distribution, the estimated covariance matrix together with the state estimates. We observe that the estimated covariance matrix is consistent with the observed error compared to the ground truth.

\section{GNSS based Localization}
\label{section:gnss}

%Position and velocity derived from GNSS are employed first to initialize and align the inertial navigation system (INS), and then as measurement with its covariance to update the filter of our framework (in section~\ref{section:fusion}). 

The RTK algorithm is implemented to fully utilize the properties of other sensors. Here we present how the RTK module is aided by our sensor fusion framework, but not the details of the implementation of RTK itself. Without loss of generality, GNSS single differenced (SD) pseudo-range and phase observations between rover ($\boldsymbol{\mathit{r}}$) and base ($\boldsymbol{\mathit{b}}$) on satellite $\boldsymbol{\mathit{i}}$ of system $\boldsymbol{\mathit{s}}$ can be constructed as:
\begin{footnotesize}
	\begin{equation} \label{equ:gnss_sd_obs}
	\begin{split}
	&\Delta\rho_{r,b}^{s,i}  = \Delta R_{r,b}^{s,i} + \Delta l_{r,b}^{s,i} \cdot dx_r + C \cdot dt_{r,b}  + \epsilon_\rho \\
	&\Delta\varphi_{r,b}^{s,i} = \Delta R_{r,b}^{s,i} + \Delta l_{r,b}^{s,i} \cdot dx_{r} + C \cdot dt_{r,b} - \lambda^{s,i}\cdot\Delta N_{r,b}^{s,i} + \epsilon_\varphi,
	\end{split}
	\end{equation}
\end{footnotesize}
where $\Delta$ represents single differenced calculating, and $\rho$ is pseudo-range while $\varphi$ for phase in meters, and $N$ stands for ambiguities to be resolved, with $\lambda$ being the wavelength of satellite $\boldsymbol{\mathit{i}}$'s certain band. $R$ is the geometry distance and $\l$ for observing matrix, both computed from satellite position with a prior rover position $x_r^0$, whose estimated correction is $dx_r$. $t$ represents relative receiver clock offsets to be estimated, with $C$ equaling the light speed. $\epsilon_\varphi$, $\epsilon_\rho$, are noises on phase and range, respectively.

%Currently, GPS, BeiDou and GLONASS share plenty of satellites in operation, and are providing, or will provide, service all over the world. 
To improve the ambiguity resolution success rate, we use all GPS, BeiDou and GLONASS observations currently available, and the GLONASS inter-frequency bias is estimated \cite{GLONASS-IFB}. With least squares, we can get SD float ambiguities and their covariance, then apply a transformation matrix to convert them into double differenced (DD) ambiguities with integer nature, after which, MLAMBDA \cite{MLAMBDA} is utilized to resolve ambiguities.

Although the ambiguity resolved (AR) RTK solution is preferred, there are indeed many situations with severe multipath and signal blockage when it is difficult to resolve ambiguities, for example, under urban buildings or in forests, where only a floated ambiguity RTK solution or code-based differential GNSS are available with sub-meter accuracy. Overall in our framework, the GNSS positioning result, either AR-RTK or float RTK, is used to update the Kalman filter with the corresponding uncertainty $\delta_g$, computed from:
\begin{footnotesize}
	\begin{equation} \label{equ:gnss_uncertainty}
	\begin {split}
	&\delta_g^2  = (B^T P B)^{-1} \cdot \frac{V^T P V}{n -r},
	\end{split}
	\end{equation}
\end{footnotesize}
where $V$ and $P$ respectively represent posterior residuals and weight matrix for SD observations, of which the number is $n$ and $r$ is the number of estimated states in Equation \ref{equ:gnss_sd_obs}, derived from which, $B$ as observing matrix with its $i$ row $B_i$ = [$\Delta l^{i}$ $C$].

\vspace{-0.1cm}
\subsection{INS-aided Ambiguity Resolution}\label{section:gnss_ar}
Without the aid of other sensors, the success rate of ambiguity resolution would heavily depend on pseudo-range precision and under urban buildings with serve multi-path, the rate may degrade greatly. However in our work, INS constrained by LiDAR (in section~\ref{section:lidar}) and/or GNSS, can provide a promising prediction to narrow down the ambiguity search space. For example, when passing through a tunnel without GNSS signals, the framework could continue to work with LiDAR constraining INS errors and provide an accurate position prediction to help resolve ambiguities until GNSS signals are reacquired.

Currently, our work only loosely couples the sensor observations (tightly coupling is planned for the future), and here we simply take the INS integration result $x_{ins}$ as a virtual observation with its variance $R_{ins}$: $\hat{x}_r = x_{ins}$.
%\vspace{-0.2cm}
%\subsection{Uncertainty Estimation}
%\crl{Add brief description of the uncertainty estimation of GNSS method here.}

\vspace{-0.1cm}
\subsection{Phase Cycle Slip Detecting}
When the GNSS receiver loses its lock on signal tracking, a sudden jump of carrier phase measurements, called cycle slip, can happen, which then forces discontinuity of integer ambiguity and worsens the positioning \cite{Takasu2008_cycle_slip}. Furthermore, under urban environments where GNSS signals are commonly obstructed and reflected, cycle slips can occur much more frequently than in static open conditions and should be detected and repaired to improve RTK performance for a mobile vehicle.

Here we estimate the rover's position incremental offset and receiver clock drift between two consecutive epochs(say $k$ and $k-1$), at 5Hz, based on consideration of the satellite geometry, with  tropospheric and ionospheric delays remaining unchanged. Time differenced observations can be written as below:
\begin{footnotesize}
	\begin{equation} \label{equ:slip_sd_obs_range_phase}
	\begin {split}
	&\Delta\rho_{k,k-1}^{s,i}  = \Delta R_{k,k-1}^{s,i} + \Delta l_{k,k-1}^{s,i} \cdot dx_{k} + C \cdot dt_{k,k-1}  + \epsilon_\rho \\
	&\Delta\varphi_{k,k-1}^{s,i} = \Delta R_{k,k-1}^{s,i} + \Delta l_{k,k-1}^{s,i} \cdot dx_{k} + C \cdot dt_{k,k-1} \\
	&- \lambda^{s,i}\cdot\Delta N_{k,k-1}^{s,i}  + \epsilon_\varphi.
	\end {split}
	\end{equation}
\end{footnotesize}
The SD phase ambiguities are estimated with SD pseudo-range. Again, the transformed DD float ambiguities together with their covariance are used to obtain fixed ambiguities with MLAMBDA engine. Obviously, a nonzero element in fixed DD ambiguities vector indicates a cycle slip on a related carrier phase that deserves a new ambiguity resolution process in section~\ref{section:gnss_ar}.

\section{Sensor Fusion}
\label{section:fusion}

In our fusion framework, an error-state Kalman filter is applied to fuse the localization measurements, discussed in the above section, with IMU. The fusion framework can optimally combine the orientation rate and accelerometer information from IMU, for improved accuracy. IMU is sufficiently accurate to provide robust state estimates between LiDAR and RTK measurements.

\vspace{-0.2cm}
\subsection{SINS Kinematics Equation and Error Equation}

A Strap-down Inertial Navigation System (SINS) estimates the position, velocity and attitude by integrating the IMU data. In this paper, we choose east-north-up (ENU) as the navigation reference frame ($n$), and right-forward-up (RFU) as the body frame ($b$) \cite{gao2015ins}, and we also use the earth frame ($e$) and the inertial frame ($i$) \cite{savage1998strapdown}.
Primarily, the differential equation \cite{gao2015ins}  \cite{savage1998strapdown}  \cite{groves2013principles} in $n$ frame of SINS is well known as:
\begin{small}
	\begin{equation} \label{equ:sins_equation}
	\begin {split}
	&    \dot{\boldsymbol{\mathit{v}}}^n = \boldsymbol{\mathit{C}}_b^n (\boldsymbol{\mathit{f}}^b - \boldsymbol{\mathit{b}}_a) - (2\boldsymbol{\mathit{\omega}}_{ie}^n + \boldsymbol{\mathit{\omega}}_{en}^n) \times \boldsymbol{\mathit{v}}^n + \boldsymbol{\mathit{g}}^n\\
	&    \dot{\boldsymbol{\mathit{r}}} = \boldsymbol{\mathit{R}}_c \boldsymbol{\mathit{v}}^n\\
	&   \dot{\boldsymbol{\mathit{q}}}_b^n = \frac{1}{2} * (\boldsymbol{\mathit{\omega}}_{nb}^b \times) \otimes \boldsymbol{\mathit{q}}_b^n,\\ 
	\end{split}
	\end{equation}
\end{small}

where
$\boldsymbol{\mathit{r}}=(\lambda,L,a)^T$ is the vehicle position; 
$\boldsymbol{\mathit{v}}^n$ is the vehicle velocity; 
$\boldsymbol{\mathit{q}}_b^n$ is the attitude quaternion from $b$ frame to $n$ frame; 
$\boldsymbol{\mathit{C}}_b^n$ is the direction cosine matrix from $b$ frame to $n$ frame; 
$\boldsymbol{\mathit{g}}^n$ is the gravity; 
$\boldsymbol{\mathit{b}}_g$ is the gyroscopes biases; $\boldsymbol{\mathit{b}}_a$ is the accelerometers biases; 
$\boldsymbol{\mathit{\omega}}_{ib}^b$, $\boldsymbol{\mathit{f}}^b$ are the IMU gyroscopes and accelerometers output respectively; 
$\boldsymbol{\mathit{\omega}}_{\alpha \beta}^\gamma$ is the angular rate of $\beta$ frame with respect to $\alpha$ frame, resolved in $\gamma$ frame;
$\otimes$ is the quaternion multiplication operator; 
$\boldsymbol{\mathit{R}}_c$ transforms the integration of velocity to longitude $\lambda$, latitude $L$ and altitude $a$, and 
$\boldsymbol{\mathit{R}}_c = diag(\frac{1}{(R_N + a)cos(L)}, \frac{1}{R_M + a}, 1)$, 
where $R_N$, $R_M$ are the transverse radius and the meridian radius, respectively.

For the tactical grade MEMS IMU, the IMU biases can be modeled as a constant value model.
When we combine the SINS with other supplementary sensors, the RTK or LiDAR, using an error-state Kalman filter, a SINS error model is necessary. The $\psi$ angle model error equation for the velocity, position and attitude error can be expressed as follows in the navigation frame  \cite{benson1975comparison}:
\begin{small}
	\begin{equation} \label{equ:sins_vel_error}
	\begin {split}
	&    \delta \dot{\boldsymbol{\mathit{v}}}^n = \boldsymbol{\mathit{C}}_b^n \boldsymbol{\mathit{f}}^b \times \delta \boldsymbol{\mathit{\psi}} - (2\boldsymbol{\mathit{\omega}}_{ie}^n + \boldsymbol{\mathit{\omega}}_{en}^n) \times \delta \boldsymbol{\mathit{v}}^n + \boldsymbol{\mathit{C}}_b^n \delta \boldsymbol{\mathit{f}}^b\\
	&    \delta \dot{\boldsymbol{\mathit{r}}} = -\boldsymbol{\mathit{\omega}}_{en}^n \times \delta \boldsymbol{\mathit{r}} + \boldsymbol{\mathit{R}}_c \delta \boldsymbol{\mathit{v}}^n\\
	&    \delta \dot{\boldsymbol{\mathit{\psi}}} = -\boldsymbol{\mathit{\omega}}_{in}^n \times \delta \boldsymbol{\mathit{\psi}} - \boldsymbol{\mathit{C}}_b^n \delta \boldsymbol{\mathit{\omega}}_{ib}^b,
	\end {split}
	\end{equation}
\end{small}
where $\delta \boldsymbol{\mathit{v}}^n$, $\delta \boldsymbol{\mathit{r}}$, $\delta \boldsymbol{\mathit{f}}^b$, $\delta \boldsymbol{\mathit{\omega}}_{ib}^b$ are the error of $\boldsymbol{\mathit{v}}^n$, $\boldsymbol{\mathit{r}}$, $\boldsymbol{\mathit{f}}^b$, $\boldsymbol{\mathit{\omega}}_{ib}^b$ respectively; $\delta \boldsymbol{\mathit{\psi}}$ is the error of attitude angle.

\vspace{-0.2cm}
\subsection{Filter State Equation}
Because the SINS error grows over time, to get a precise PVA, we use an error-state Kalman filter to estimate the error of the SINS and use the estimated error-state to correct the SINS. 

Especially for $\boldsymbol{\mathit{q}}_b^n$ in Equation \ref{equ:sins_equation}, the perturbation quaternion $\delta \boldsymbol{\mathit{q}}_b^n$ can be expressed as a small angle approximation when it is assumed to be a very small angle around $0$:
\begin{small}
	\begin{equation} \label{equ:quaternion_map}
	\delta \boldsymbol{\mathit{q}}_b^n 
	= \exp (\begin{bmatrix} 0 \\ \frac{1}{2}\delta \boldsymbol{\mathit{\psi}} \end{bmatrix})
	= \begin{bmatrix} \cos(\left \| \frac{\delta \boldsymbol{\mathit{\psi}}}{2} \right \|) \\ \sin(\left \| \frac{\delta \boldsymbol{\mathit{\psi}}}{2} \right \|) \frac {\delta \boldsymbol{\mathit{\psi}}} {\| \delta \boldsymbol{\mathit{\psi}} \|} \end{bmatrix}
	\approx (\begin{bmatrix} 1 \\ \frac{1}{2}\delta \boldsymbol{\mathit{\psi}} \end{bmatrix}).
	\end{equation}
\end{small}

Therefore, we choose the state variables as $\boldsymbol{\mathit{X}} = \begin{bmatrix} \boldsymbol{\mathit{r}} & \boldsymbol{\mathit{v}}^n & \boldsymbol{\mathit{q}}_b^n & \boldsymbol{\mathit{b}}_a & \boldsymbol{\mathit{b}}_g \end{bmatrix}^T$, and the state variables' error as $\delta \boldsymbol{\mathit{X}} = \begin{bmatrix} \delta \boldsymbol{\mathit{r}} & \delta \boldsymbol{\mathit{v}}^n & \delta \boldsymbol{\mathit{\psi}} & \delta \boldsymbol{\mathit{b}}_a & \delta \boldsymbol{\mathit{b}}_g \end{bmatrix}^T$.

From the SINS error Equation (\ref{equ:sins_vel_error}) and the IMU model, we can obtain the state equation of the Kalman filter as follows:
\begin{small}
	\begin{equation} \label{equ:kalman_state_equ}
	\delta \dot{\boldsymbol{\mathit{X}}} = \boldsymbol{\mathit{F(\boldsymbol{\mathit{X}})}} \delta \boldsymbol{\mathit{X}}
	+ \boldsymbol{\mathit{G(\boldsymbol{\mathit{X}})} \boldsymbol{\mathit{W}}},
	\end{equation}
\end{small}
where $\boldsymbol{\mathit{W}} = \begin{bmatrix} \boldsymbol{\mathit{w}}_a & \boldsymbol{\mathit{w}}_g & \boldsymbol{\mathit{w}}_{b_a} & \boldsymbol{\mathit{w}}_{b_g} \end{bmatrix}^T$ is the system noise, which comprises the IMU output noise and the IMU bias noise.

$\left ( \cdot \times \right )$ denotes the skew-symmetric matrix of a vector. $\boldsymbol{\mathit{F(\boldsymbol{\mathit{X}})}}$ and $\boldsymbol{\mathit{G(\boldsymbol{\mathit{X}})}}$ can be expressed as the following equations:\\
$
\begin{scriptsize}
\boldsymbol{\mathit{F(\boldsymbol{\mathit{X}})}} =
\begin{bmatrix}
-(\mathbf{\mathit{\omega_{en}^n}} \times) & \boldsymbol{\mathit{R}}_c & \mathbf{0}_{3\times 3} & \mathbf{0}_{3\times 3} & \mathbf{0}_{3\times 3}\\ 
\mathbf{0}_{3\times 3} & -((2\boldsymbol{\mathit{\omega}}_{ie}^n + \boldsymbol{\mathit{\omega}}_{en}^n) \times) & ((\boldsymbol{\mathit{C}}_b^n \boldsymbol{\mathit{f}}^b) \times) & \mathbf{\mathit{C_{b}^n}} & \mathbf{0}_{3\times 3}\\ 
\mathbf{0}_{3\times 3} & \mathbf{0}_{3\times 3} & -(\mathbf{\mathit{\omega_{in}^n}} \times) & \mathbf{0}_{3\times 3} & -\mathbf{\mathit{C_{b}^n}}\\ 
\mathbf{0}_{6\times 3} & \mathbf{0}_{6\times 3} & \mathbf{0}_{6\times 3} & \mathbf{0}_{6\times 3} & \mathbf{0}_{6\times 3}
\end{bmatrix}\\
\end{scriptsize}
$
$
\begin{scriptsize}
\boldsymbol{\mathit{G(\boldsymbol{\mathit{X}})}} = 
\begin{bmatrix}
\mathbf{0}_{3\times 3} & \mathbf{0}_{3\times 3} & \mathbf{0}_{3\times 6}\\ 
\mathbf{\mathit{C_{b}^n}} & \mathbf{0}_{3\times 3} & \mathbf{0}_{3\times 6}\\ 
\mathbf{0}_{3\times 3} & -\mathbf{\mathit{C_{b}^n}} & \mathbf{0}_{3\times 6}\\ 
\mathbf{0}_{6\times 3} & \mathbf{0}_{6\times 3} & \mathbf{I}_{6\times 6}
\end{bmatrix}.
\end{scriptsize}
$

\vspace{-0.2cm}
\subsection{Filter Measurement Update Equation}
The measurement update comprises the LiDAR and GNSS parts. The measurement update step of the error-state Kalman filter updates the uncertainty of the state given a global correction $\boldsymbol{\mathit{Z}}$. The rest of the time or measurement update just follows the standard Kalman filter.

\subsubsection{LiDAR Measurement Update Equation}
LiDAR based localization outputs the position and heading angle of the vehicle as the filter measurement. The measurement update equation can be expressed as follows:
\begin{footnotesize}
	\begin{equation} \label{equ:lidar_measure_equ}
	\begin{split}
	%\boldsymbol{\mathit{Z}}_{L} &= \begin{bmatrix} \lambda_{S} & L_{S} & a_{S} & h_{S} \end{bmatrix}^T - \begin{bmatrix} \lambda_{L} & L_{L} & a_{L} & h_{L} \end{bmatrix}^T\\
	\boldsymbol{\mathit{Z}}_{L} &= (\lambda_{S}, L_{S}, a_{S}, h_{S})^T - (\lambda_{L}, L_{L}, a_{L}, h_{L})^T\\
	% \boldsymbol{\mathit{Z}}_{LiDAR} &= h_{LiDAR}(\boldsymbol{\mathit{X}}) + \boldsymbol{\mathit{V}}_{LiDAR}\\
	&= \boldsymbol{\mathit{H}}_{L} \delta \boldsymbol{\mathit{X}} + \boldsymbol{\mathit {V}}_{L},
	\end{split}
	\end{equation}
\end{footnotesize}
where the variable with $S$ subscript is the prediction from SINS and the variable with $L$ subscript is the LiDAR measurement. $\boldsymbol{\mathit {V}}_{L}$ is the estimation noise of the LiDAR with zero mean and its covariance matrix is $\boldsymbol{\mathit {R}}_{L}$. From Equation (\ref{equ:histogram2}), we compute the corresponding part of $\lambda$ and $L$ in $\boldsymbol{\mathit {R}}_{L}$. We set the part of $a$ and $h$ constant.
% $0.03^2$ and $0.15rad^2$; 
$\boldsymbol{\mathit{H}}_{L}$ can be obtained from following equation:
\begin{footnotesize}
	\begin{equation} \label{equ:LiDAR_measure_h}
	\boldsymbol{\mathit{H}}_{L} = \begin{bmatrix}
	\mathbf{I}_{3 \times 3} & \mathbf{0}_{3 \times 3} & \mathbf{0}_{3 \times 1} &  \mathbf{0}_{3 \times 1} & \mathbf{0}_{3 \times 1} & \mathbf{0}_{3 \times 6}\\ 
	\mathbf{0}_{1 \times 3} & \mathbf{0}_{1 \times 3} & \frac{-c_{12}c_{32}}{c_{22}^2 + c_{12}^2} & \frac{-c_{22}c_{32}}{c_{22}^2 + c_{12}^2} & 1 & \mathbf{0}_{3 \times 6}
	\end{bmatrix},
	% \end{split}
	\end{equation}
\end{footnotesize}
where $c_{ij}$ is the $i$ row and $j$ column element of $\boldsymbol{\mathit{C}}_b^n$. 

\subsubsection{GNSS Measurement Update Equation}
GNSS can estimate the position of the vehicle. The measurement update equation can be expressed as follows:
\begin{scriptsize}
	\begin{equation} \label{equ:gnss_measure_equ}
	\begin {split}
	%\boldsymbol{\mathit{Z}}_{G} &= \begin{bmatrix} \lambda_{S} & L_{S} & a_{S} \end{bmatrix}^T - \begin{bmatrix} \lambda_{G} & L_{G} & a_{G} \end{bmatrix}^T\\
	\boldsymbol{\mathit{Z}}_{G} &= (\lambda_{S}, L_{S}, a_{S})^T - (\lambda_{G}, L_{G}, a_{G})^T\\
	&    = \boldsymbol{\mathit{H}}_{G} \delta \boldsymbol{\mathit{X}} + \boldsymbol{\mathit{V}}_{G},
	\end {split}
	\end{equation}
\end{scriptsize}
where the variable with $G$ subscript is the GNSS measurement. $\boldsymbol{\mathit{V}}_{G}$ is the estimation noise of the GNSS with zero mean and its covariance matrix is $\boldsymbol{\mathit {R}}_{G}$ obtained from Equation (\ref{equ:gnss_uncertainty}). $\boldsymbol{\mathit{H}}_{G}$ is defined as: $\boldsymbol{\mathit{H}}_{G} = \begin {bmatrix} \mathbf{I}_{3 \times 3} & \mathbf{0}_{3 \times 12} \end {bmatrix}$.

\vspace{-0.2cm}
\subsection{Delay Handling}
\begin{figure}[htbp]
	\centering
	\includegraphics[scale=0.35]{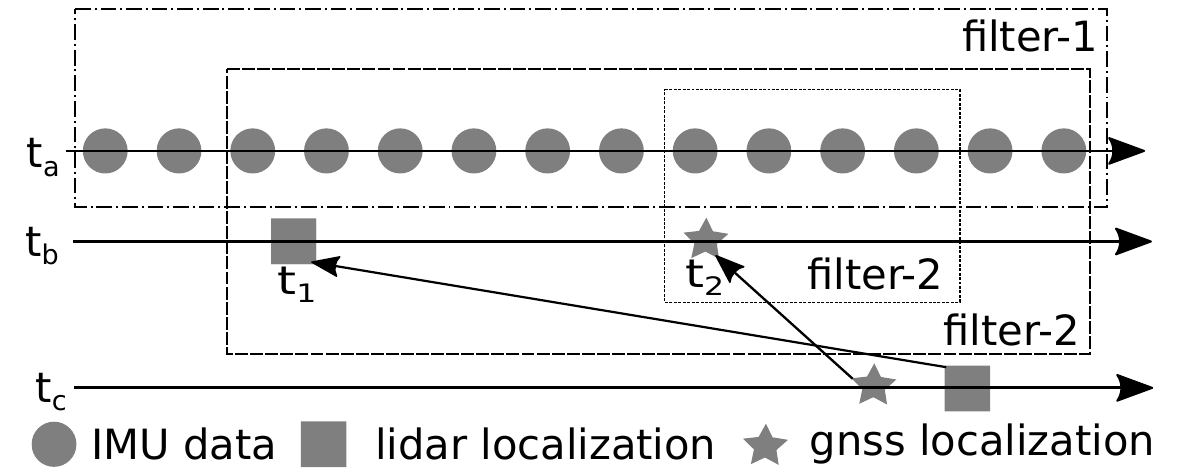}
	\caption{\footnotesize Delay and disorder measurement processing. $t_a$, $t_b$ and $t_c$ represent the IMU time series, the measurement occurred time series, and the measurement received time series, respectively. The rectangle and the star represent two measurements that received in the wrong order.} 
	\label{fig:sensor-delay}
	\vspace{-0.2cm}
\end{figure}

Due to the transmission and computation delay, the measurement delay and disorder must be taken into consideration. The solution is that we maintain two filters and a fixed length buffer of the filter states, the filter measurements and the IMU data, in chronological order. The filter-1 computes the real-time PVA and its covariance by performing the time update and integrating the IMU data instantly when new IMU data is received. The filter-2 processes the delayed measurement. When a new measurement at $t_1$ is received, we execute the following actions:
\begin{enumerate}
	\item Obtain the filter states at $t_1$ from the buffer. Update the filter states in filter-2.
	\item Execute the measurement update at $t_1$ in filter-2.
	\item Execute the time update in filter-2 using the IMU data in the buffer until it reaches the current time. Or we stop at $t_2$, if another measurement is found at $t_2$ in the buffer, where $t_2$ is later than $t_1$. This measurements at $t_1$ and $t_2$ are received in the wrong order.
	\item Execute the measurement update at $t_2$, if there is another measurement at $t_2$. Then repeat Step 3 and find more measurements received in the wrong order.
	\item When we finish the time update and reach the current time, the filter states are in the buffer and the states of the filter-1 are updated according to the new results starting from $t_1$ to the current time.
\end{enumerate}
Figure \ref{fig:sensor-delay} is used to illustrate this procedure.

\section{Experimental Results}
\label{section:experiments}
Our testing platform is shown in Fig \ref{fig:intro}.
%We can easily replace the integrated navigation system, NovAtel ProPak6 plus NovAtel IMU-IGM-A1, with a more compact and integrated product at much lower price (\$5,000 - \$10,000). NovAtel IMU-IGM-A1 is a tactical grade MEMS IMU. We plan to test our system with lower grade IMU in the future.
Ground-truth vehicle motion trajectories are generated using offline methods for quantitative analysis. In open spaces with good GNSS signal reception, the GNSS/INS solution based on post-processing algorithms, such as the NovAtel Inertial Explorer, is able to produce enough accurate vehicle motion trajectories. In weak GNSS signal scenarios, such as complex urban roads, we treat it as a classic map reconstruction problem combining several techniques including NovAtel IE post-processing, LiDAR SLAM, loop closure, and the global pose-graph optimization. We only show qualitative results in demo video clips for GNSS-denied scenes, such as tunnel or underground garage. Thus, we  classify our testing datasets (60km in total) into three general categories: 
\begin{enumerate*}
	\item 48.1km regular roads: YF-1, YF-2, YF-3, YF-4, YF-5 covering common road conditions, such as urban, countryside, and a traffic jam.
	\item 10.4km weak GNSS signal roads: HBY-1, DS-1 covering narrow roads lined with tall buildings or trees.
	\item 2.1km GNSS-denied roads: DT-1 covering tunnels.
\end{enumerate*}

%\begin{itemize}
%\item Regular roads: We collect 5 logs of data on regular roads. The total length of this dataset is about 48.1 km, covering common road conditions, such as urban, rural, traffic congestion, and roadside parking. The dataset consists of YF-1, YF-2, YF-3, YF-4, YF-5.
%\item Weak-GPS roads: We collect 2 logs of data on poor-GPS roads. One where there are tall buildings near the road, and another where the road is covered by trees. The total length of this dataset is 10.4 km. The two datasets consist of HBY-1, DS-1.
%\item Non-GPS roads: We collect a log containing a 1.2 km underground bridge where the GPS signal would completely disappear. The length of this dataset is 4.5 km, containing 2.4 km underground bridge and 2.1 km regular road. 
%\end{itemize}

\vspace{-0.2cm}
\subsection{Quantitative Analysis}
\begin{figure*}[htbp]
	\centering
	%\subfloat[YF-1]{
	%	\begin{minipage}{9.0cm}
	%		\centering
	%		\includegraphics[width=\textwidth]{figures/yongfeng1.pdf}
	%	\end{minipage}
	%}
	%\subfloat[YF-4]{
	%	\begin{minipage}{9.0cm}
	%		\centering
	%		\includegraphics[width=\textwidth]{figures/yongfeng4.pdf}
	%	\end{minipage}
	%} \\
	%\subfloat[DS-1]{
	%	\begin{minipage}{9.0cm}
	%		\centering
	%		\includegraphics[width=\textwidth]{figures/dasha.pdf}
	%	\end{minipage}
	%} 	
	%\subfloat[HBY-1]{
	%	\begin{minipage}{9.0cm}
	%		\centering
	%		\includegraphics[width=\textwidth]{figures/huanbaoyuan.pdf}
	%	\end{minipage}
	%}	
	\subfloat[Longitudinal RMS of YF-1]{
		\begin{minipage}{8.7cm}
			\centering
			\includegraphics[width=\textwidth]{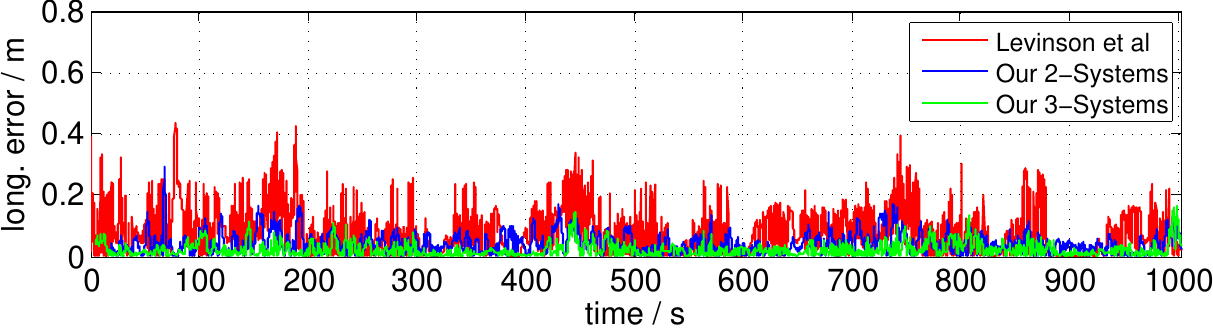}
		\end{minipage}
	}
	\subfloat[Lateral RMS of YF-1]{
		\begin{minipage}{8.7cm}
			\centering
			\includegraphics[width=\textwidth]{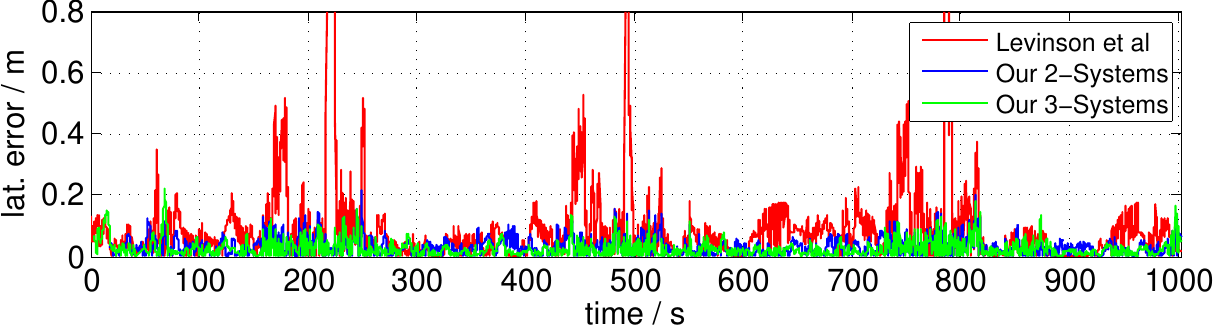}
		\end{minipage}
	}
	\caption{\footnotesize Quantitative comparison with \cite{Levinson_etal2010}. Our proposed solution achieves better performance over \cite{Levinson_etal2010} consistently over time.} 
	\label{fig:exp1a}
	\vspace{-0.2cm}
\end{figure*}

Our system has been extensively tested in real-world driving scenarios. We compare our localization performance against the state-of-the-art intensity-based localization method proposed by Levinson et al. \cite{Levinson_etal2007, Levinson_etal2010}. In order to explicitly demonstrate the contribution of different sensors, the test results are shown in two modes: 
\begin{enumerate*}
	\item 2-Systems: LiDAR + IMU
	\item 3-Systems: LiDAR + GNSS + IMU. 
\end{enumerate*}
In Table \ref{table:tab1}, we show the quantitative results in both regular or weak GNSS roads. Note our vast performance improvement over \cite{Levinson_etal2010} and the robust and accurate localization results in both regular and weak GNSS scenarios with centimeter level accuracy. That both the 2-Systems and 3-Systems work well demonstrates that our system does not rely on a single sensor but fuses the sensors input using resilient and adaptive methods. We display the lateral and longitudinal error over time in Fig. \ref{fig:exp1a}. As exhibited, our proposed solution is able to achieve better performance over \cite{Levinson_etal2010} consistently over time.

\vspace{-0.2cm}
\subsection{Qualitative Analysis}
For GNSS-denied roads, we do not present quantitative comparisons due to the lack of ground truth. In our additional video clips, we show the qualitative comparison between our results and NovAtel's GNSS RTK/IMU poses. The reason we did not show the result of \cite{Levinson_etal2010} is that it fails when NovAtel's RTK/IMU poses are not stable and smooth enough.

\begin{figure*}[htbp]
	\centering
	\subfloat[]{
		\begin{minipage}[t]{0.132\linewidth}
			\centering
			\includegraphics[width=1.0\textwidth]{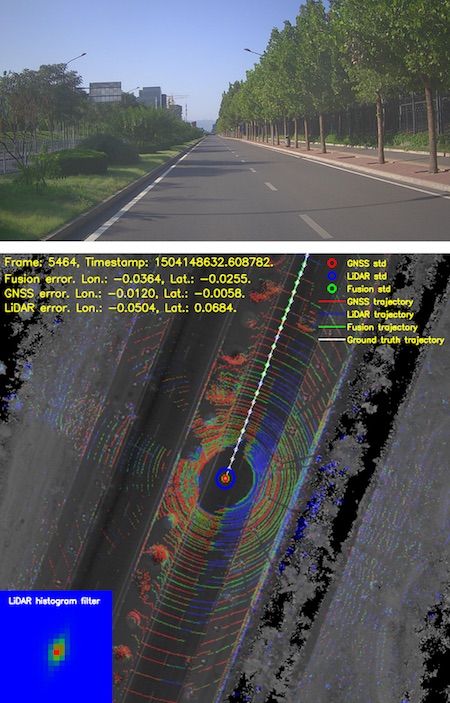}
		\end{minipage}
	}
	\subfloat[]{
		\begin{minipage}[t]{0.132\linewidth}
			\centering
			\includegraphics[width=1.0\textwidth]{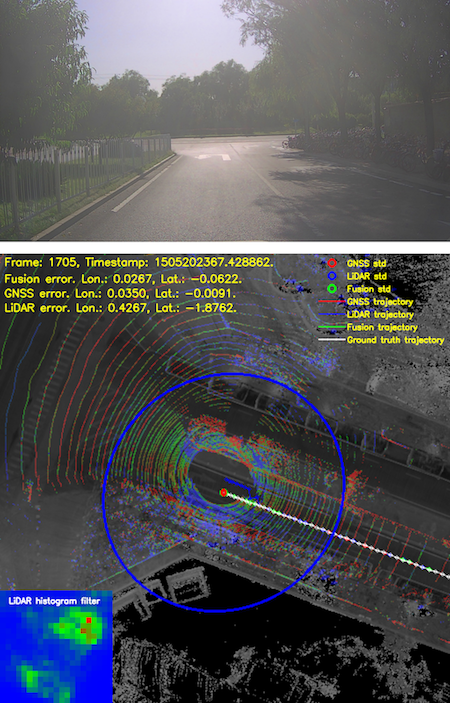} 
		\end{minipage}
	}		
	\subfloat[]{
		\begin{minipage}[t]{0.132\linewidth}
			\centering
			\includegraphics[width=1.0\textwidth]{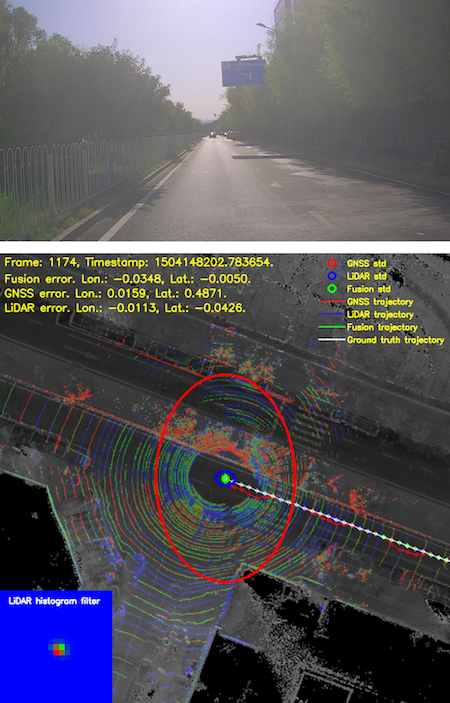} 
		\end{minipage}
	}
	\subfloat[]{
		\begin{minipage}[t]{0.132\linewidth}
			\centering
			\includegraphics[width=1.0\textwidth]{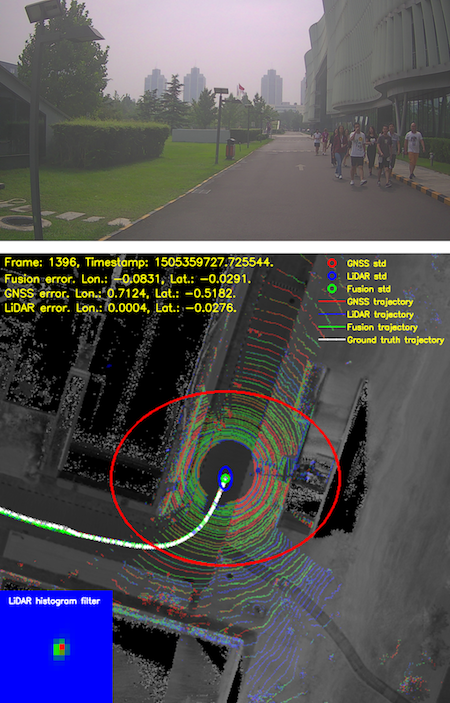} 
		\end{minipage}
	}
	\subfloat[]{
		\begin{minipage}[t]{0.132\linewidth}
			\centering
			\includegraphics[width=1.0\textwidth]{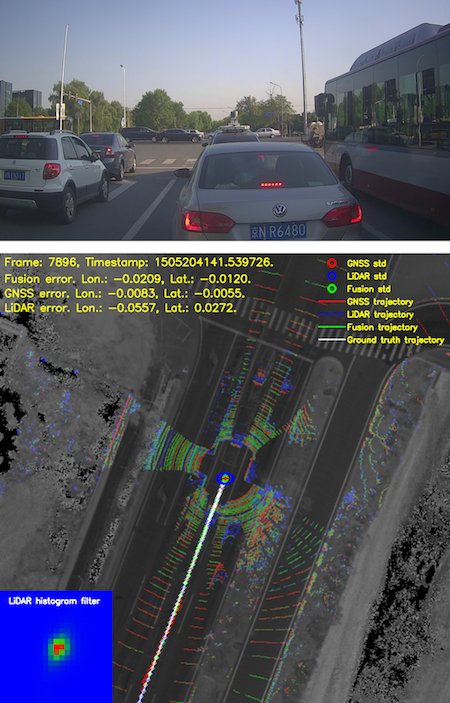} 
		\end{minipage}
	}
	\subfloat[]{
		\begin{minipage}[t]{0.267\linewidth}
			\centering
			\includegraphics[width=1.0\textwidth]{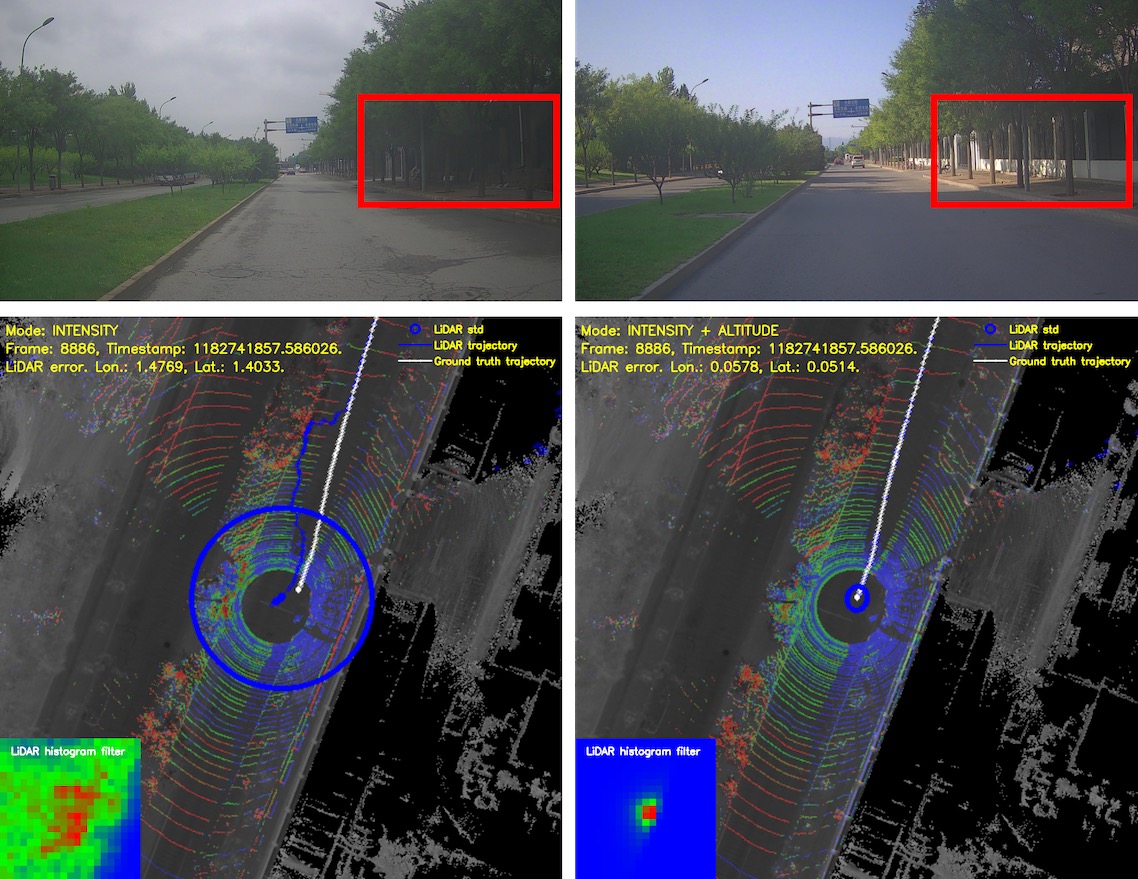} 
		\end{minipage}
	}
	\caption{\footnotesize Performance analysis of each individual function. The first row is the front camera image of the scene. The figures in the second row indicate the performance of our system. Green, blue and red ellipses represent the uncertainty of the localization estimation of each method: Fusion, LiDAR and RTK, respectively. The small blue figure in the corner represents the posterior density in the histogram filter of the LiDAR module. Figures (a) - (f) show the performance of our system under different circumstances. (a) Both the LiDAR and RTK give good results. (b) The LiDAR gives poor results due to an outdated map. (c) The RTK gives unstable results due to the signal blockage. (d) - (e) The system performs well with crowds or cars around. (f) shows that the LiDAR fails when only the intensity cue is used but succeeds when we make use of both the intensity and altitude cues on a newly paved road with a recently built wall.} 
	\label{fig:exp1b}
	\vspace{-0.2cm}
\end{figure*}

In Figure~\ref{fig:exp1b}, we give performance analysis of each module and function in our system in detail. (a) shows a typical case where both the RTK and LiDAR give good results, as does the fused system. In (b), the LiDAR fails due to an outdated map. However, the fused system gives excellent results with the aid of the RTK. We show an opposite example in (c) where the RTK is poor due the signal blockage, and the LiDAR is in good working condition. (d) and (e) demonstrate the good performance of our system in crowded scenes with people or cars. (f) shows a very interesting case with a newly paved road and a recently built wall. The LiDAR cannot handle such significant environmental changes based only on the intensity cues. It gives good results when we adaptively fuse the additional altitude cues.

\begin{table}[htbp]
	\centering  
	\begin{tabular}{l|llllll}  
		\toprule[1pt]
		Logs & \makecell[tl]{Method} & \makecell[tl]{Horiz. \\ RMS} & \makecell[tl]{Horiz. \\ Max} & \makecell[tl]{Long. \\ RMS} & \makecell[tl]{Lat. \\ RMS} & \makecell[tl]{$<$ 0.3m \\ Pct.} \\  
		\midrule[.5pt] 
		\multirow{2}{*}{\makecell[tl]{Regular \\ Roads}} & \cite{Levinson_etal2010} & 0.209 & 1.934 & 0.097 & 0.161 & 82.25\% \\
		& \cellcolor{mygray}2-Sys & \cellcolor{mygray}0.075 & \cellcolor{mygray}0.560 & \cellcolor{mygray}0.050 & \cellcolor{mygray}0.045 & \cellcolor{mygray}99.42\% \\
		& \cellcolor{mygray}3-Sys & \cellcolor{mygray}0.054 & \cellcolor{mygray}0.551 & \cellcolor{mygray}0.032 & \cellcolor{mygray}0.036 & \cellcolor{mygray}99.54\% \\
		\midrule[.5pt] 
		\multirow{2}{*}{\makecell[tl]{Weak-GNSS \\ Roads}} & \cite{Levinson_etal2010} & 0.143 & 0.737 & 0.088 & 0.093 & 95.02\% \\
		& \cellcolor{mygray}2-Sys & \cellcolor{mygray}0.070 & \cellcolor{mygray}0.315 & \cellcolor{mygray}0.050 & \cellcolor{mygray}0.039 & \cellcolor{mygray}99.99\% \\
		& \cellcolor{mygray}3-Sys & \cellcolor{mygray}0.073 & \cellcolor{mygray}0.258 & \cellcolor{mygray}0.053 & \cellcolor{mygray}0.041 & \cellcolor{mygray}100.0\% \\
		\bottomrule[1pt] 
	\end{tabular}
	\caption{\footnotesize Quantitative comparison with \cite{Levinson_etal2010}. The performance of two modes of our system is shown: 1) 2-Systems: LiDAR + IMU; 2) 3-Systems: LiDAR + GNSS + IMU. The benefits of GNSS in regular roads are clearly visible. Our localization error is far lower than \cite{Levinson_etal2010}.}
	\label{table:tab1}
\end{table}

\vspace{-0.2cm}
\subsection{Detailed Analysis of LiDAR}
\label{section:lidarexperiments}
To demonstrate the effectiveness of each of the above contributions in Section \ref{section:lidar}, we show the localization accuracy with different methods in Table \ref{table:tab3} using the 2-Systems. We also introduce a special data log (YF-6), which includes a district where the road was newly paved after the map data collection. {\tt Intensity} denotes the baseline method where only intensity cues are used. In {\tt Heading}, we add the heading estimation step, which is especially helpful for the low-grade IMU. In {\tt FixedAlt}, we incorporate the altitude cues during the LiDAR matching. Note the large improvement that clearly demonstrates the effectiveness of the altitude cues. Further, in {\tt AdaptAlt}, we incorporate the altitude cues with adaptive weights. From the result of YF-4 in table~\ref{table:tab3}, we observe that our heading angle optimization step is crucial to LiDAR localization. The result of YF-6 in table~\ref{table:tab3} indicates that our adaptive weighting strategy makes the system more robust to environmental changes, such as road construction, and seasonal variations. Demonstrably, this gives us the lowest localization errors in all metrics.

\begin{table}[htbp]
	\centering  
	\begin{tabular}{l|llllll}  
		\toprule[1pt]
		Logs & \makecell[tl]{Method} & \makecell[tl]{Horiz. \\ RMS} & \makecell[tl]{Horiz. \\ Max} & \makecell[tl]{Long. \\ RMS} & \makecell[tl]{Lat. \\ RMS} & \makecell[tl]{$<$ 0.3m \\ Pct.} \\  
		\midrule[.5pt] 
		\multirow{3}{*}{YF-4} & Intensity & 0.153 & 1.090 & 0.107 & 0.085 & 94.41\% \\
		& Heading & 0.081 & 0.461 & 0.060 & 0.042 & 99.63\% \\
		& FixedAlt & 0.074 & 0.341 & 0.054 & 0.040 & 99.97\% \\
		& \cellcolor{mygray}AdaptAlt & \cellcolor{mygray}0.068 & \cellcolor{mygray}0.307 & \cellcolor{mygray}0.048 & \cellcolor{mygray}0.039 & \cellcolor{mygray}99.98\% \\
		\midrule[.5pt] 
		\multirow{3}{*}{YF-6} & Intensity & 0.508 & 11.928 & 0.368 & 0.293 & 92.72\% \\ 
		& Heading & 0.485 & 12.771 & 0.352 & 0.282 & 95.68\% \\
		& FixedAlt & 0.216 & 8.016 & 0.151 & 0.129 & 95.89\% \\
		& \cellcolor{mygray}AdaptAlt & \cellcolor{mygray}0.057 & \cellcolor{mygray}0.319 & \cellcolor{mygray}0.038 & \cellcolor{mygray}0.035 & \cellcolor{mygray}99.99\% \\
		\bottomrule[1pt] 
	\end{tabular}
	\caption{\footnotesize Comparison of localization errors of various methods used in our system. The benefits of each of heading angle refinement, altitude cues and adaptive weights are clearly visible.}
	\label{table:tab3}
\end{table}

% Yong Feng
%\begin{table}[h]
%\centering    	
%   	\begin{tabular}{l|llllll}  
%   		\toprule[1pt]
%   		Method & Mean & Std & Max & $<$ 0.3m & $<$ 0.2m & $<$ 0.1m  \\  
%   		\midrule[.5pt] 
%   		Intensity & 0.508 & 2.131 & 11.928 & 92.72\% & 90.46\% & 77.51\% \\     		
%   		Heading & 0.485 & 2.150 & 12.771 & 95.68\% & 95.24\% & 90.17\% \\
%   		FixedAlt & 0.216 & 0.856 & 8.016  & 95.89\% & 95.48\% & 90.71\% \\
%   		\rowcolor{mygray} AdaptAlt & 0.051 & 0.032 & 0.373  & 99.92\% & 99.59\% & 92.22\% \\
%   		\bottomrule[1pt] 
%   	\end{tabular}  
%   	\caption{YF-6 registration result}  
%   	\label{table:tab3}
%\end{table}  

\vspace{-0.2cm}
\subsection{Run-time Analysis}
Our system primarily contains three modules: LiDAR, GNSS, and SINS. LiDAR, GNSS, and SINS work at 10hz, 5hz, and 200hz, respectively. There are two implementations designed for different applications. One occupies only a single CPU core. The other uses a single CPU core plus a FPGA. In the single core version, we reduce the computation load by using a smaller histogram filter size and  downsampling data during the heading angle evaluation. Both versions provide similar localization results in terms of accuracy, but a larger histogram filter size can potentially increase the possibility of the convergence when the filter drifts away in an abnormal event. GNSS and SINS modules only take about 0.2 CPU core.

%For GNSS-component, it takes about 5ms, on average, to finish its one observation computing.

%For SINS-component, it is based on dual-core CPU, one for filter-1 and one for filter-2 in figure ~\ref{fig:sensor-delay}. Filter-1 processes IMU data real-time, and can get the vehicle pose in less than 1ms. Filter-2 processes LiDAR localization and GNSS localization, with a computing time of less than 10ms when it receives a GNSS localization and 10ms-20ms when it receives a LiDAR localization.

\section{Conclusion and Future Work}
We have presented a complete localization system, designed for fully autonomous driving applications. It adaptively fuses the input from complementary sensors, such as GNSS, LiDAR and IMU, to achieve good localization accuracy in various challenging scenes, including urban downtown, highways or expressways, and tunnels. Our system achieves 5-10cm RMS accuracy both longitudinally and laterally and is ready for industrial use by having two versions with different computing hardware requirements. Our system, deployed in a large autonomous driving fleet, makes our vehicles fully autonomous in crowded city streets every day. The generality of our fusion framework means it can be used to readily fuse more sensors at various cost levels, facing different applications. Actually, we have begun testing our system with low-cost, low-end MEMS IMU. Our future work also includes building a low-cost localization solution designed for ADAS or Level 3 self-driving car.

%\addtolength{\textheight}{-12cm}   % This command serves to balance the column lengths
                                  % on the last page of the document manually. It shortens
                                  % the textheight of the last page by a suitable amount.
                                  % This command does not take effect until the next page
                                  % so it should come on the page before the last. Make
                                  % sure that you do not shorten the textheight too much.

%%%%%%%%%%%%%%%%%%%%%%%%%%%%%%%%%%%%%%%%%%%%%%%%%%%%%%%%%%%%%%%%%%%%%%%%%%%%%%%%

%%%%%%%%%%%%%%%%%%%%%%%%%%%%%%%%%%%%%%%%%%%%%%%%%%%%%%%%%%%%%%%%%%%%%%%%%%%%%%%%

%%%%%%%%%%%%%%%%%%%%%%%%%%%%%%%%%%%%%%%%%%%%%%%%%%%%%%%%%%%%%%%%%%%%%%%%%%%%%%%%

%\section*{APPENDIX}
%Appendixes should appear before the acknowledgment.

\begin{small}
\section*{ACKNOWLEDGMENT}
We would like to thank our colleagues for their kind help and support throughout the project. Weixin Lu helped with the vehicle preparation. Yao Zhou and Cheng Wang helped with the demo video production. Shichun Yi, Li Yu and Cheng Wang generated the LiDAR map. Nadya Bosch helped with the text editing.
\end{small}

\bibliographystyle{IEEEtran}
\bibliography{IEEEabrv,localization}

\end{document}